\documentclass[10pt,twocolumn,letterpaper]{article}

\usepackage{cvpr}              
\usepackage{color}
\usepackage{comment}

\usepackage{algorithm}
\usepackage{algpseudocode}   
\usepackage{bbm}

\usepackage{tabularray}
\usepackage{float}

\newcommand{\method}[1]{{NewtPhys}}

\definecolor{darkmagenta}{RGB}{186, 0, 186}
\definecolor{slateblue}{RGB}{90, 70, 230}

\definecolor{todo}{HTML}{E15554}
\definecolor{author1}{HTML}{40A050}
\definecolor{author2}{HTML}{C33C54}
\definecolor{author3}{HTML}{E6AA68}
\definecolor{author4}{HTML}{0F4893}

\definecolor{material}{HTML}{2BAE27}
\definecolor{mechanics}{HTML}{FF5733}
\definecolor{spatial}{HTML}{3498DB}
\definecolor{permanence}{HTML}{F43FC7}
\definecolor{temporal}{HTML}{0DA792}
\definecolor{viewpoint}{HTML}{EEAC32}

\makeatletter
\DeclareRobustCommand\onedot{\futurelet\@let@token\@onedot}
\def\@onedot{\ifx\@let@token.\else.\null\fi\xspace}

\definecolor{best}{HTML}{C6EFCE}    
\definecolor{second}{HTML}{FFF2CC}  

\definecolor{pgrade1}{RGB}{255,204,204}  
\definecolor{pgrade2}{RGB}{255,221,204}  
\definecolor{pgrade3}{RGB}{255,238,204}  
\definecolor{pgrade4}{RGB}{255,255,204}  
\definecolor{pgrade5}{RGB}{229,255,204}  
\definecolor{pgrade6}{RGB}{204,255,204}  
\definecolor{pgrade7}{RGB}{178,242,187}  
\definecolor{pgrade8}{RGB}{153,230,170}  
\definecolor{pgrade9}{RGB}{102,204,136}  
\definecolor{pgrade10}{RGB}{51,153,102}  

\definecolor{corr0}{RGB}{255,255,255}  
\definecolor{corr1}{RGB}{229,245,233}
\definecolor{corr2}{RGB}{204,232,207}
\definecolor{corr3}{RGB}{168,221,181}
\definecolor{corr4}{RGB}{123,204,196}
\definecolor{corr5}{RGB}{78,179,211}
\definecolor{corr6}{RGB}{43,140,190}
\definecolor{corr7}{RGB}{29,116,155}
\definecolor{corr8}{RGB}{20,90,120}
\definecolor{corr9}{RGB}{0,68,89}      

\definecolor{cvprblue}{rgb}{0.21,0.49,0.74}
\usepackage[pagebackref,breaklinks,colorlinks,allcolors=cvprblue]{hyperref}

\def\paperID{18}
\def\confName{CVPR}
\def\confYear{2026}

\title{Training-Free Fine-Grained Semantic Segmentations in Low Data Regimes: \\A FungiTastic Baseline}

\author{
Sebastian Cavada \quad Francesco Pelosin \quad Lapo Faggi\\
Covision Lab, Bressanone, South Tyrol, Italy\\
\texttt{\{name.surname\}@covisionlab.com}
}

\begin{document}
\twocolumn[{%
\renewcommand\twocolumn[1][]{#1}%
\maketitle
\begin{center}
\begin{minipage}[t]{0.49\textwidth}
    \centering
    \includegraphics[width=\linewidth]{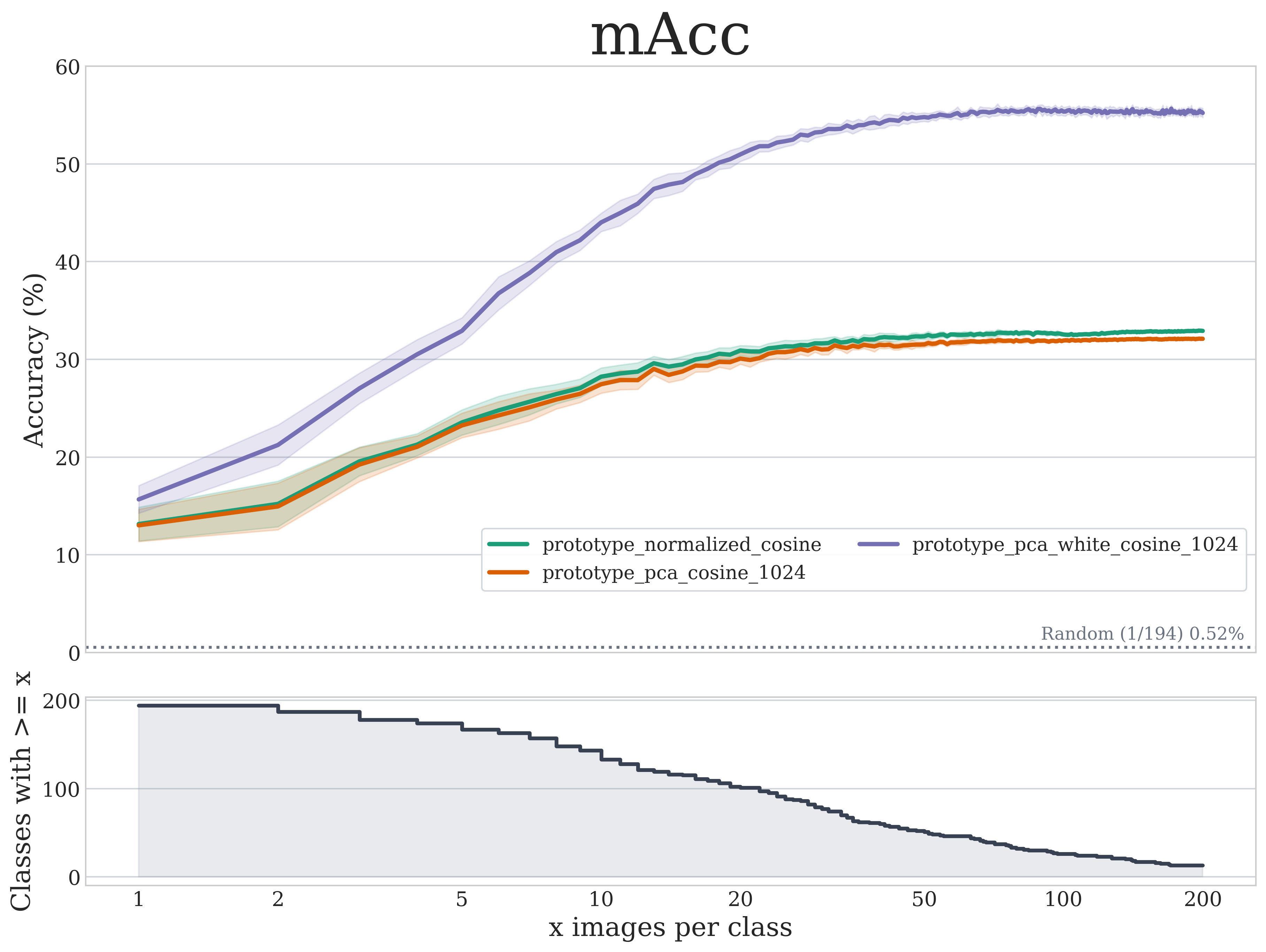}

    \vspace{0.15em}
\end{minipage}
\hfill
\begin{minipage}[t]{0.49\textwidth}
    \centering
    \includegraphics[width=\linewidth]{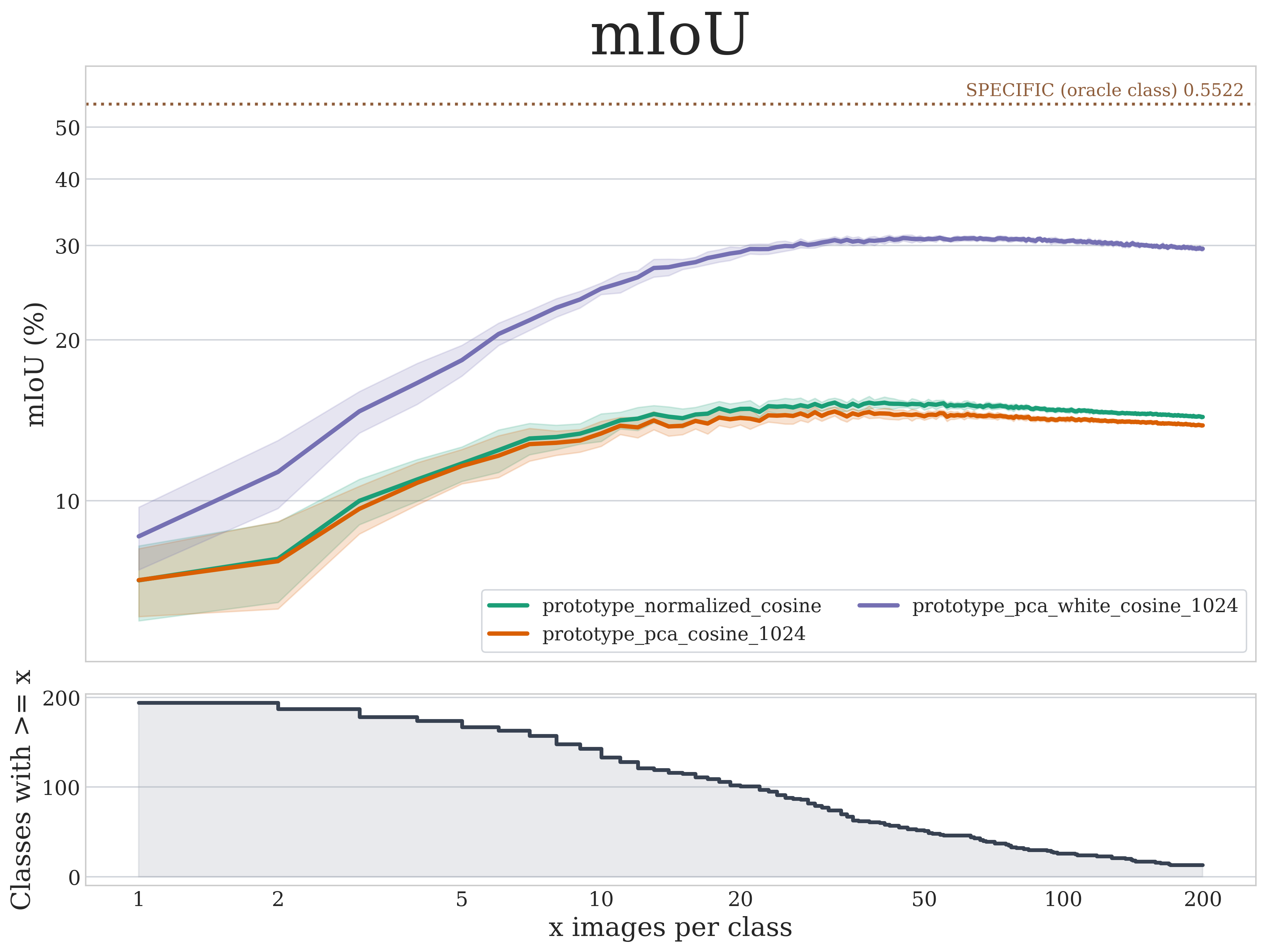}

    \vspace{0.15em}
\end{minipage}

\vspace{0.4em}

\captionsetup{type=figure}
\caption{
Performance across low-data regimes on FungiTastic. Top: mean class accuracy (left) and mean IoU (right) as the number of training images per class increases. PCA whitening consistently improves prototype matching over raw normalized DINOv3 features and vanilla PCA, yielding the best accuracy and segmentation performance. Performance saturates after roughly 40--60 images per class, suggesting that a relatively small subset is sufficient to cover most of the dataset variability. Bottom: number of classes with at least $x$ training images, highlighting the long-tailed nature of the benchmark.}
\label{fig:teaser}
\end{center}

\vspace{0.6em}
}]
\begin{abstract}

Fine-grained semantic segmentation requires both precise localization and discrimination between visually similar classes. In FungiTastic, this problem is further complicated by a long-tailed distribution and strong variation in image acquisition conditions.

We propose a training-free two-stage framework that decouples segmentation from classification. SAM3 first produces class-agnostic mushroom masks using macro-taxonomic prompts, and DINOv3 then assigns fine-grained labels through prototype matching in the embedding space. To improve this stage, we apply a simple transformation of the DINOv3 feature space that improves prototype-based classification.

Compared with class-specific prompting, our approach is more scalable and keeps the segmentation cost low. We report results from one-shot to few-hundred-shot regimes, providing, to the best of our knowledge, the first baseline for fine-grained semantic segmentation in low-data settings.

\end{abstract}
\vspace{-1em}

\section{Introduction}

Fine-grained semantic segmentation requires both accurate localization and the ability to distinguish subtle visual differences between classes \cite{birdclef, plantclef}. This challenge is particularly severe in mycology, where species often exhibit high visual similarity, strong intra-class variability, and long-tailed distributions. The FungiTastic dataset \cite{Picek_2022_WACV} further increases the difficulty due to large variations in the acquisition conditions.

In practice, progress in this setting is also limited by the cost of dense pixel-level annotation, which makes fine-grained segmentation especially challenging under low-data regimes. This is a critical issue in specialized domains, where annotated data are scarce and difficult to collect.

Recent vision foundation models \cite{dinov2, clip, siglip} provide strong off-the-shelf representations and make low-training or training-free pipelines increasingly attractive. At the same time, SAM3 \cite{carion2025sam3segmentconcepts} has shown strong class-agnostic segmentation capabilities, but its semantic use still depends on class-specific prompting. In fine-grained settings, this is problematic, since the correct class prompt is not known at inference time, and exhaustive prompting scales linearly with the number of classes.

Therefore, we assume only class labeling without any segmentation supervision. To do so, we first obtain class-agnostic mushroom masks with SAM3 using a macro-taxonomic prompt, then assign fine-grained labels by matching DINOv3 \cite{simeoni2025dinov3} features to class prototypes. This removes the need for class-conditioned segmentation at inference time, keeping the segmentation cost constant with respect to the label space while remaining suitable for low-data regimes.
\vspace{-1em}

\paragraph{Contributions}
\begin{itemize}
    \item We propose a simple two-stage framework that decouples segmentation and classification for fine-grained semantic segmentation.
    \item We show that transformed DINOv3 embeddings significantly improve prototype-based matching.
    \item We establish a benchmark for fine-grained semantic segmentation in low-data regimes.
\end{itemize}

\begin{figure*}
    \centering
    \includegraphics[width=1\linewidth]{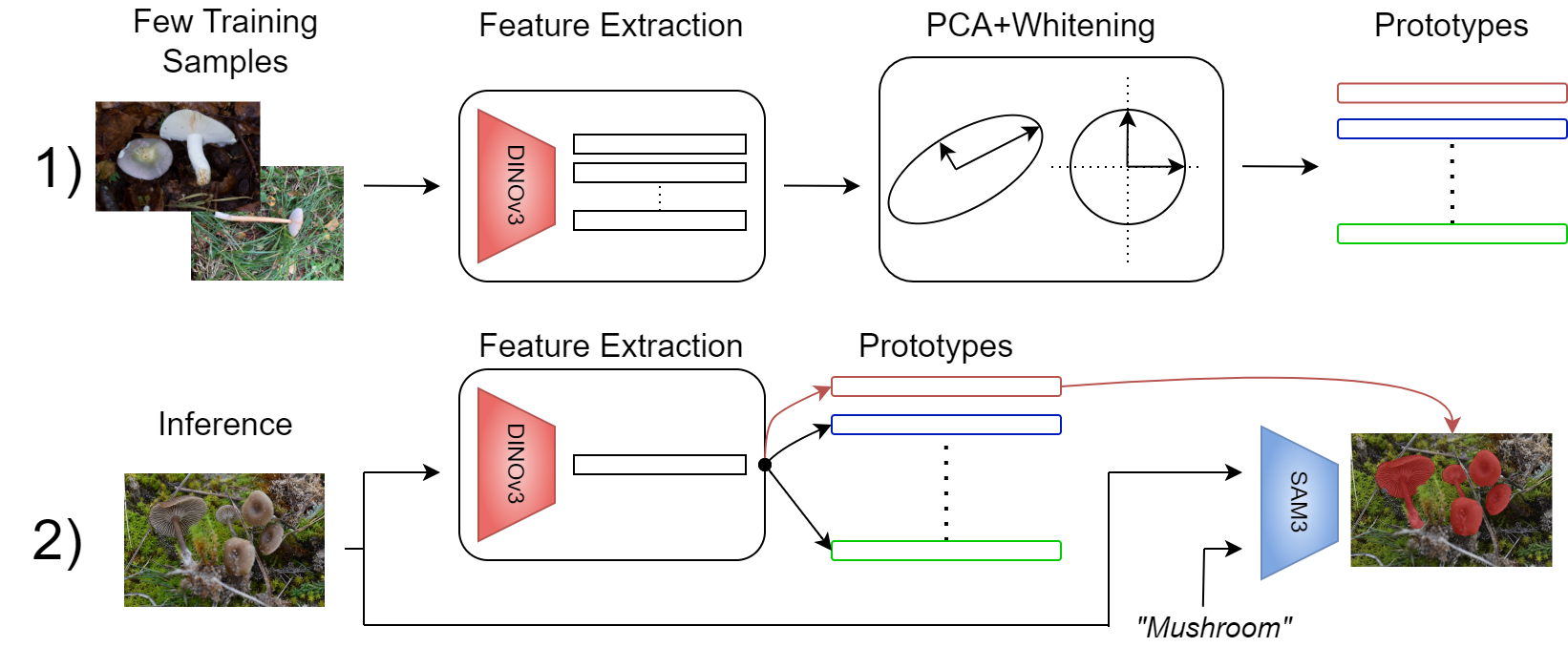}
    \caption{
Macro-to-fine pipeline: DINOv3 extracts features to match class prototypes for label prediction, while SAM3 provides class-agnostic segmentation. The predicted label is assigned to the segmented region to obtain fine-grained masks.
}
    \label{fig:pipeline}
\end{figure*}

\section{Related Works}

\paragraph{Fine-grained learning in low-data regimes.}
Fine-grained learning under limited supervision has been studied mainly for classification. The survey in \cite{tang2024awesome_fgfs} shows that prior work is largely restricted to image-level recognition, with little attention to dense prediction. Fine-grained semantic segmentation is more demanding, as it requires both recognition and precise localization. Only a small number of recent works, such as \cite{part_matching}, address related settings, mainly through part-level correspondence rather than category-level segmentation.

\vspace{-1em}
\paragraph{FungiTastic benchmark and prior solutions.}
FungiTastic \cite{Picek_2025_CVPR} has so far been explored mostly for classification. Recent FungiCLEF submissions, such as \cite{prototypical_networks, contrastive}, focus on long-tailed recognition through feature engineering, ensembling, and lightweight finetuning. These methods target classification and do not address fine-grained segmentation.

\vspace{-1em}
\paragraph{Prototype-based methods.}
Prototype-based learning is a standard approach for few-shot recognition \cite{proto1, proto2, proto3}, where classes are represented in an embedding space. Recent works also use multimodal prototypes to improve generalization under class imbalance \cite{multimodal}. Our method follows this idea, but combines prototype-based classification with class-agnostic segmentation in a single pipeline.

\vspace{-1em}
\paragraph{PCA whitening and feature preprocessing} PCA whitening is a standard preprocessing technique for distance-based inference in embedding spaces, as it decorrelates features and rescales principal directions to unit variance \cite{kessy2018optimal}. We suggest that such transformations are especially useful when pretrained representations are dominated by high-variance directions that may reflect nuisance factors more than task-discriminative structure \cite{joint}. Recent work has also shown that whitening can improve the quality of visual self-supervised representations \cite{weng2022investigation, kalapos2024whitening}, further supporting its relevance in representation-based recognition pipelines.
\section{Method and Experimental Setup}
\label{sec:setup}

\begin{table}[t]
\centering
\resizebox{\columnwidth}{!}{%
\begin{tabular}{ccccc}
\textbf{Method} & \textbf{mIoU} & \textbf{Non-empty/Total Images} \\
\hline
Macro-Taxonomy & 0.8937 & 9643/9763 \\
Fine-Grained (Oracle) & 0.5522 & 6341/9763 \\

\end{tabular}%
}
\caption{mIoU comparison of prompting strategies on the test set. The first row uses generic prompting followed by oracle fine-grained classification. The second row assumes oracle fine-grained classification first and then applies fine-grained prompting.}
\label{tab:macromicro}
\end{table}

\subsection{Dataset}
We evaluate our approach in the FungiTastic dataset \cite{Picek_2025_CVPR}, a fine-grained benchmark characterized by subtle differences between classes and a highly imbalanced, long-tailed distribution. We consider only the portion of the dataset comprising segmentation masks. The subset contains approximately 13k images with segmentation annotations for training set, and 9k in the test set, spanning over 194 different classes. We report results in the one-shot, few-shot, and low-data regimes as the mean and standard deviation of the evaluation metrics over 20 runs on the test set.

\subsection{Evaluation Protocol}

\paragraph{General Baseline + Oracle.} We prompt SAM3 with the string ``mushrooms" which constitutes our macro taxonomic prompt and retain the output by thresholding at 0.5. This constitutes the general ability of the model to segment all kinds of mushrooms with class-agnostic knowledge. We then assume an oracle classification of pixels in order to compute the macro taxonomic upper bound as shown in Table~\ref{tab:macromicro} (i.e., how good is SAM3 at segmenting mushrooms).

\vspace{-1em}
\paragraph{Fine-Grained Baseline + Oracle.} To produce fine-grained SAM segmentations, we cannot assume a priori class knowledge; therefore, we would need to forward each test image 194 times (i.e., one for each class) and devise a post-hoc procedure to select the most probable class. This results in a segmentation cost that grows linearly with the number of classes. To this end, we assume that an oracle provides exact taxonomic scientific names, and we use these as prompts for SAM3. We then retain the output by thresholding to 0.3, adopting a lower threshold than in the macro-taxonomic baseline, to account for the reduced confidence scores we observed when prompting SAM3 with specific classes. This baseline is useful to investigate the abilities of SAM3 when prompted with exact classes (ex. ``boletus edulis" or ``amanita muscaria"). This also acts as our upper bound, and the result is reported in Table~\ref{tab:macromicro}.

\vspace{-1em}
\paragraph{Our Method}
Our approach decomposes the task into a class-agnostic segmentation stage followed by a training-free fine-grained classification stage. We first obtain macro-taxonomic segmentations using SAM3 prompted with a generic concept (``mushrooms''), ensuring that segmentation complexity remains constant with respect to the number of classes.

For representation learning, we employ DINOv3 as a frozen backbone. Unless otherwise specified, we use the \texttt{[CLS]} token as the image-level feature representation. To simulate low-data regimes, we construct multiple training subsets by uniformly sampling from the available training data and repeat experiments across $n=20$ different random seeds.

During the prototype construction phase, DINOv3 features are extracted for each training sample and subsequently projected using PCA followed by whitening. This transformation improves the geometry of the feature space, facilitating more reliable distance-based comparisons. Class prototypes are then computed as the mean of normalized features within each class.

At inference time, given a test image, we extract its feature representation using DINOv3 and assign a class label by selecting the nearest prototype in the transformed feature space. In parallel, the same image is processed by SAM3 to obtain a macro-level segmentation mask. The final fine-grained segmentation is produced by propagating the predicted class label to all pixels belonging to the macro segmentation mask, effectively converting class-agnostic regions into class-specific predictions. An explanation in Figure~\ref{fig:pipeline} reports the full pipeline.

We note that the order of the two stages could be reversed, with classification performed first and fine-grained prompting to SAM3 applied afterward. However, as shown in Table~\ref{tab:macromicro}, performing segmentation first and classification second yields superior accuracy, and is therefore adopted in our pipeline.

\begin{table*}[ht]
\centering
\small
\begin{tabular}{lcccccc|cccccc}
 & \multicolumn{6}{c}{\textbf{mAcc}} & \multicolumn{6}{c}{\textbf{mIoU}} \\
\cmidrule(lr){2-7} \cmidrule(lr){8-13}
 & k=5 & k=10 & k=20 & k=50 & k=100 & k=200 & k=5 & k=10 & k=20 & k=50 & k=100 & k=200 \\
\midrule
Norm. cosine & $0.24$ & $0.28$ & $0.31$ & $0.32$ & $0.33$ & $0.33$ & $0.12$ & $0.14$ & $0.15$ & $0.15$ & $0.15$ & $0.14$ \\
PCA cosine & $0.23$ & $0.27$ & $0.30$ & $0.32$ & $0.32$ & $0.32$ & $0.12$ & $0.13$ & $0.14$ & $0.14$ & $0.14$ & $0.14$ \\
PCA white cosine & $0.33$ & $0.44$ & $0.51$ & $0.55$ & $0.55$ & $0.55$ & $0.18$ & $0.25$ & $0.29$ & $0.31$ & $0.31$ & $0.30$ \\
\bottomrule
\end{tabular}
\vspace{2mm} 
\caption{Prototype-based classification results averaged over 20 seeds for different numbers of samples per class $k$. The maximum standard deviation is at most 0.01.}
\label{tab:prototype_results_mean_std}
\end{table*}

\subsection{Metrics.}
We report mean class accuracy (mAcc) for image-level classification and mean Intersection over Union (mIoU) for segmentation. Let $C$ denote the number of classes, and let $\mathbf{M^{I}} \in \mathbb{N}^{C \times C}$ be the image-level confusion matrix and $\mathbf{M^{P}} \in \mathbb{N}^{C \times C}$ the pixel-level confusion matrix. We consider $M_{ij}$ as the number of samples whose ground-truth class is $i$ and predicted class is $j$.

\vspace{-1em}
\paragraph{Mean Accuracy  (mAcc)}
\begin{equation}
\mathrm{mAcc} = \frac{1}{C} \sum_{i=1}^{C} \frac{M^{I}_{ii}}{\sum_{j=1}^{C} M^{I}_{ij}},
\end{equation}
that is, the average across classes of the diagonal entries normalized by the total number of ground-truth samples in each class.

\vspace{-1em}
\paragraph{Mean IoU (mIoU)}
\begin{equation}
\mathrm{mIoU} = \frac{1}{C} \sum_{i=1}^{C} \frac{M^{P}_{ii}}{\sum_{j=1}^{C} M^{P}_{ij} + \sum_{j=1}^{C} M^{P}_{ji} - M^{P}_{ii}},
\end{equation}
which corresponds to the average, over all classes, of the ratio between true positives and the union of ground-truth and predicted pixels for that class.

\section{Results}
\label{sec:results}

\paragraph{Accuracy} Results are reported in tabular form in Table \ref{tab:prototype_results_mean_std} and visually shown in Figure~\ref{fig:teaser} (left) show that standard normalized DINOv3 embeddings without any feature preprocessing achieve a performance $\sim 30\%$ of mAcc. Taking into account feature preprocessing unlocks more than $\sim 20\%$ of improvement, reaching mAcc $\sim 50\%$.  This peculiar result is discussed in the following paragraph.

\vspace{-1.5em}
\paragraph{Feature Preprocessing}
FungiTastic contains substantial nuisance variability, such as changes in background, illumination, zoom, and broader acquisition conditions. These factors strongly affect the geometry of DINOv3 CLS-token embeddings and can obscure directions that are more relevant for fine-grained fungi classification. As shown in Figure~\ref{fig:teaser} (right), neither raw-feature normalization in the original $4096$-dimensional space nor vanilla PCA is sufficient to separate classes well. This points to a mismatch between the structure of the pretrained embedding space and the metric needed for prototype comparison. PCA whitening alleviates this issue: after projection onto the principal-component basis, each retained component is rescaled by its standard deviation. The resulting representation is more balanced across directions, which reduces the influence of dominant nuisance-related components and makes class-relevant variation easier to exploit.

\vspace{-1.5em}

\paragraph{mIoU}
In the proposed pipeline, the final segmentation performance depends on the interplay between the prototype-based classifier and the segmentation masks provided upfront by SAM3 queried with the generic macro-taxonomic class ``mushroom".  
When considering the resulting mIoU performances as shown in Figure~\ref{fig:teaser}, we can see that with a subset of 50 images, the benchmark reaches its maximum at mIoU $30\%$. This suggests that a small quantity of images is enough to provide full dataset coverage.

\section{Conclusion and Future Work}
We presented a simple yet effective training-free baseline for fine-grained semantic segmentation in low-data regimes, built on a two-stage decomposition of the problem into class-agnostic segmentation and fine-grained classification. By combining SAM3 macro-taxonomy masks with DINOv3 prototype-based recognition, our approach removes the need for class-specific prompting at inference time and yields a significantly more scalable alternative to exhaustive fine-grained querying. On FungiTastic, this design provides a practical baseline for a particularly challenging setting characterized by subtle inter-class differences, long-tailed class distributions, and strong nuisance variability across acquisition conditions.

Our results show that the main bottleneck is not only the scarcity of supervision, but also the geometry of the embedding space used for prototype matching. Off-the-shelf DINOv3 [CLS] features are not sufficiently aligned with the metric structure required for fine-grained discrimination in this setting. In contrast, correcting the feature geometry through PCA whitening leads to a marked improvement, as it suppresses nuisance-dominated high-variance directions and allows more class-informative components to contribute to cosine-based comparison. This finding suggests that, in low-data fine-grained regimes, representation preprocessing can be as important as the choice of the underlying foundation model itself.

Beyond the specific benchmark, our study highlights a broader message: effective fine-grained dense prediction baselines can emerge from carefully composed foundation models, even without task-specific training. We hope this work provides a useful reference point for future research on low-data fine-grained segmentation.
Future work will extend this framework by incorporating masked [PATCH] tokens alongside global descriptors, exploring alternative visual backbones, and evaluating the approach in multi-class segmentation settings and additional benchmarks.

Overall, we believe this work constitutes a first baseline for fine-grained semantic segmentation under realistic low-data conditions, and highlights the potential of carefully composed training-free pipelines when their design is matched to the structure of the problem.

{
    \small
    \bibliographystyle{ieeenat_fullname}
    \bibliography{main}
}

\end{document}